%% file: main.tex
\documentclass[10pt,twocolumn,letterpaper]{article}

\usepackage{cvpr} 
\usepackage{multirow}
\usepackage[accsupp]{axessibility}  

\input{preamble}

\definecolor{cvprblue}{rgb}{0.21,0.49,0.74}
\usepackage[pagebackref,breaklinks,colorlinks,allcolors=cvprblue]{hyperref}
\usepackage{bm}
\usepackage[linesnumbered,ruled]{algorithm2e}
\def\paperID{11790} 
\def\confName{CVPR}
\def\confYear{2025}
\newcommand{\gdd}[1]{\textcolor{black}{#1}}
\def\rr{\textcolor{black}}
\title{Beyond Words: Augmenting Discriminative Richness via Diffusions in Unsupervised Prompt Learning}
\author{Hairui Ren$^{1}$\hspace{10pt} Fan Tang$^{2}$ \hspace{10pt} He Zhao$^{3}$ \hspace{10pt} Zixuan Wang$^{1}$ \hspace{10pt} Dandan Guo$^{1*}$  \hspace{10pt} Yi Chang$^{1*}$\\
{$^1$} School of Artificial Intelligence, Jilin University \\ 
{$^2$} Institute of Computing Technology, Chinese Academy of Sciences\hspace{20pt} {$^3$} CSIRO's Data61  \\
{\tt\small \{renhr22,zixuan24\}@mails.jlu.edu.cn,tangfan@ict.ac.cn,he.zhao@ieee.org,\{guodandan,yichang\}@jlu.edu.cn}
}

\begin{document}
\maketitle

\input{Sections/0_abstract}
\input{Sections/1_Introduction}
\input{Sections/2_related}
\input{Sections/3_method}
\input{Sections/4_experiments}
\input{Sections/5_conclusion}
\input{Sections/7_Ack}


{
    \small
    \bibliographystyle{ieeenat_fullname}
    \bibliography{main}
}


\end{document}

%% file: preamble.tex
\usepackage{lineno}

\newcommand{\inc}[2]{$#1^{\color{red}+#2}$}


%% file: Sections/0_abstract.tex
\begin{abstract}
Fine-tuning vision-language models (VLMs) with large amounts of unlabeled data has recently garnered significant interest. 
However, a key challenge remains the lack of high-quality pseudo-labeled data. 
Current pseudo-labeling strategies often struggle with mismatches between semantic and visual information, leading to sub-optimal performance of unsupervised prompt learning (UPL) methods.
In this paper, we introduce a simple yet effective approach called \textbf{A}ugmenting D\textbf{i}scriminative \textbf{R}ichness via Diffusions (AiR), toward learning a richer discriminating way to represent the class comprehensively and thus facilitate classification.
Specifically, our approach includes a pseudo-label generation module that leverages high-fidelity synthetic samples to create an auxiliary classifier, which captures richer visual variation, bridging text-image-pair classification to a more robust image-image-pair classification. 
Additionally, we exploit the diversity of diffusion-based synthetic samples to enhance prompt learning, providing greater information for semantic-visual alignment.
Extensive experiments on five public benchmarks, including RESISC45 and Flowers102, and across three learning paradigms-UL, SSL, and TRZSL-demonstrate that AiR achieves substantial and consistent performance improvements over state-of-the-art unsupervised prompt learning methods.
\href{https://github.com/Hrren/Beyond-Words}{Code} is available.

\end{abstract}

%% file: Sections/1_Introduction.tex
\section{Introduction}
\label{sec:intro}
Large pre-trained vision-language models (VLMs)~\cite{radford2021learning, yuan2021florence, jia2021scaling} have achieved impressive performance in open-world visual concept learning without requiring task-specific training; however, they still benefit from adaptation for optimal results.
Prompt tuning \cite{ge2023domain, jia2022visual, 10740179, 10342801} offers an efficient strategy to enhance VLMs' discriminative accuracy on downstream tasks by learning inputs to the model.

While prompt learning can yield notable improvements, a large amount of labeled data is still required to enhance VLMs' performance across diverse downstream tasks. 
This adaptation requirement leads to considerable labeling costs, as labeled data is hard to obtain \cite{tanwisuth2023pouf, zhou2022learning, jia2024revealing, zhou2022conditional}. 
To address this, unsupervised prompt learning (UPL) \cite{huang2022unsupervised, menghini2023enhancing, lai2023padclip, shu2022test} has been introduced, leveraging the zero-shot capabilities of VLMs to generate pseudo labels, enabling effective use of unlabeled data.
For pseudo-labeling, traditional methods \cite{huang2022unsupervised} typically form a text classifier by inputting prompts with class names into the VLMs' text encoder. 
Then, image embeddings of unlabeled data are matched to these class-based text classifiers, assigning pseudo-labels to images with high-confidence predictions. 
Some other approaches \cite{menghini2023enhancing, zhang2024candidate} generate multiple candidate pseudo-labels by adjusting confidence thresholds or optimize the model with different numbers of pseudo-labeled samples at various training stages. 
While these methods apply distinct strategies for filtering pseudo-labels or refining prompts, most ultimately rely on pseudo-labels produced by text classifiers, making the quality of these pseudo-labels critically important.

\input{Figures/intro}

\gdd{In this work, we aim to improve the pseudo label quality in UPL, which can thus be easily combined with other UPL methods. Here, we first explore conduct a pilot experiment to empirically demonstrate the deficiency of text classifiers.  }
Specifically, we leverage the text encoder of CLIP~\cite{radford2021learning} to generate pseudo labels on the Flowers102 dataset~\cite{nilsback2008automated} and subsequently calculated the confusion matrix between the ground-truth labels and predicted class labels.
The results, as shown in Fig.~\ref{fig:intro}, reveal that a substantial number of samples are incorrectly predicted, $e.g.$, most ``Thorn apple'' flowers are misclassified as ``Giant white arum lily'', ``Tree mallow'' is misclassified as ``Mexican petunia''.
Why this happened? The discrimination of unlabeled images heavily relies on the precision of text classifiers, which are constructed using the words of prompt with category label names.
\gdd{However, when only using the text of ``a photo of a + class name'' to describe the image, using ``Giant white arum lily'' as the class name is even closer in semantics to the given image than ``Thorn apple''.
Because the Thorn apple flower appears to be a shade of white but the word ``Thorn apple'' does not express that, resulting in the misclassified samples. It motivates us to learn a richer discriminating way to represent the class comprehensively and thus facilitate classification. Since an image has rich information, we consider a novel solution: in addition to the text classifier built on the short text description, can we design another auxiliary classifier based on the most representative images for each category? However, this solution is problematic since we do not know the ground-truth label of each training image in the unsupervised prompt learning task, making it a dead-end loop.}

\gdd{To address the above problem, in this paper, we draw inspiration from the image generation literature. Recent advancements in image generation have made it possible to generate fidelity and diverse images with only text word descriptions, such as Stable Diffusion \cite{rombach2021highresolution}. It is effective in generating diverse images with richer visual appearance variation while preserving the key semantics. Surprisingly, we find that this text-to-image generation model can be naturally utilized to generate the synthetic representative images for each class in UPL tasks only with the text descriptions on the class names. Therefore, we propose Beyond Words: \textbf{A}ugmenting D\textbf{i}scriminative \textbf{R}ichness via Diffusions, a new framework \textbf{AiR} which applies diffusion models to generate synthetic images, which serve as a bridge from text-image-pair to image-image-pair classification, as a way to discriminate the unlabeled images. In our AiR framework, we adopt Stable Diffusion \cite{rombach2021highresolution} to generate discriminative richness, where we fine-tune the Stable Diffusion model with LoRA \cite{hu2022lora} by means of pseudo-labeled samples with high confidence to ensure prediction fidelity. Then, we introduce cosine similarity-based filtration to select the most representative synthetic image of each class generated by fine-tuned Diffusion, which is used to build the auxiliary classifier from the view of the image. Combining the predictions of the original text classifier and the auxiliary classifier, we can endow the training images with more accurate pseudo labels, improving the UPL performance. Besides, the diffusion-based synthetic samples can also be utilized to learn the prompt. Notably, our proposed AiR only needs to know the dataset name and class name of downstream tasks, which can be seamlessly integrated into most UPL methods}. Experimental results show that AiR achieves a notable improvement in classification accuracy by an average of 2.6\% in comparison to the state-of-the-art UPL method. To sum up, our contributions can be included:
\begin{itemize}
    \item We highlight that relying solely on the text classifier to generate pseudo labels in UPL is inadequate and propose a richer discriminating way to represent the class comprehensively and thus facilitate classification.
    \item We present a novel approach AiR: \textbf{A}ugmenting D\textbf{i}scriminative \textbf{R}ichness via Diffusions. 
    This method leverages the diversity and fidelity of synthetic samples to build an auxiliary classifier that captures richer variations in visual appearance while preserving essential semantics, bridging text-image-pair classification to image-image-pair classification.
    \item Our AiR demonstrates substantial improvement across three distinct unsupervised paradigms and five widely-used datasets, significantly outperforming the current state-of-the-art UPL method by an average margin of over 2.6\%.
\end{itemize}

\input{Figures/method}


%% file: Figures/intro.tex
\begin{figure}[!tb]
\centering
\includegraphics[width=1\linewidth]{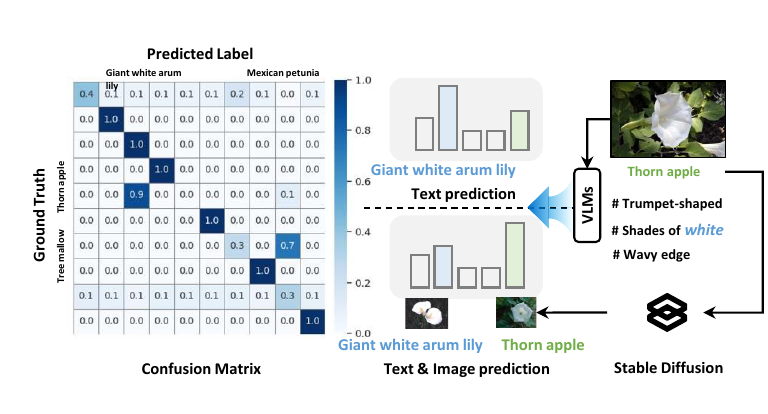}
\caption{
Left: Confusion matrix for ground truth labels $vs.$ text classifier predictions on the Flowers102 dataset, highlighting persistent generation of incorrect pseudo-labels.
Right: An example of misclassification in Flowers102, where ``Thorn apple" is misidentified as ``Giant white arum lily" due to its white petals resembling the latter’s semantic information. This misclassification can be alleviated by jointly considering text and image predictions.
}
\vspace{-20pt}
\label{fig:intro}
\end{figure}

%% file: Figures/method.tex
\begin{figure*}[!t]
\centering
\includegraphics[width=0.75\linewidth]{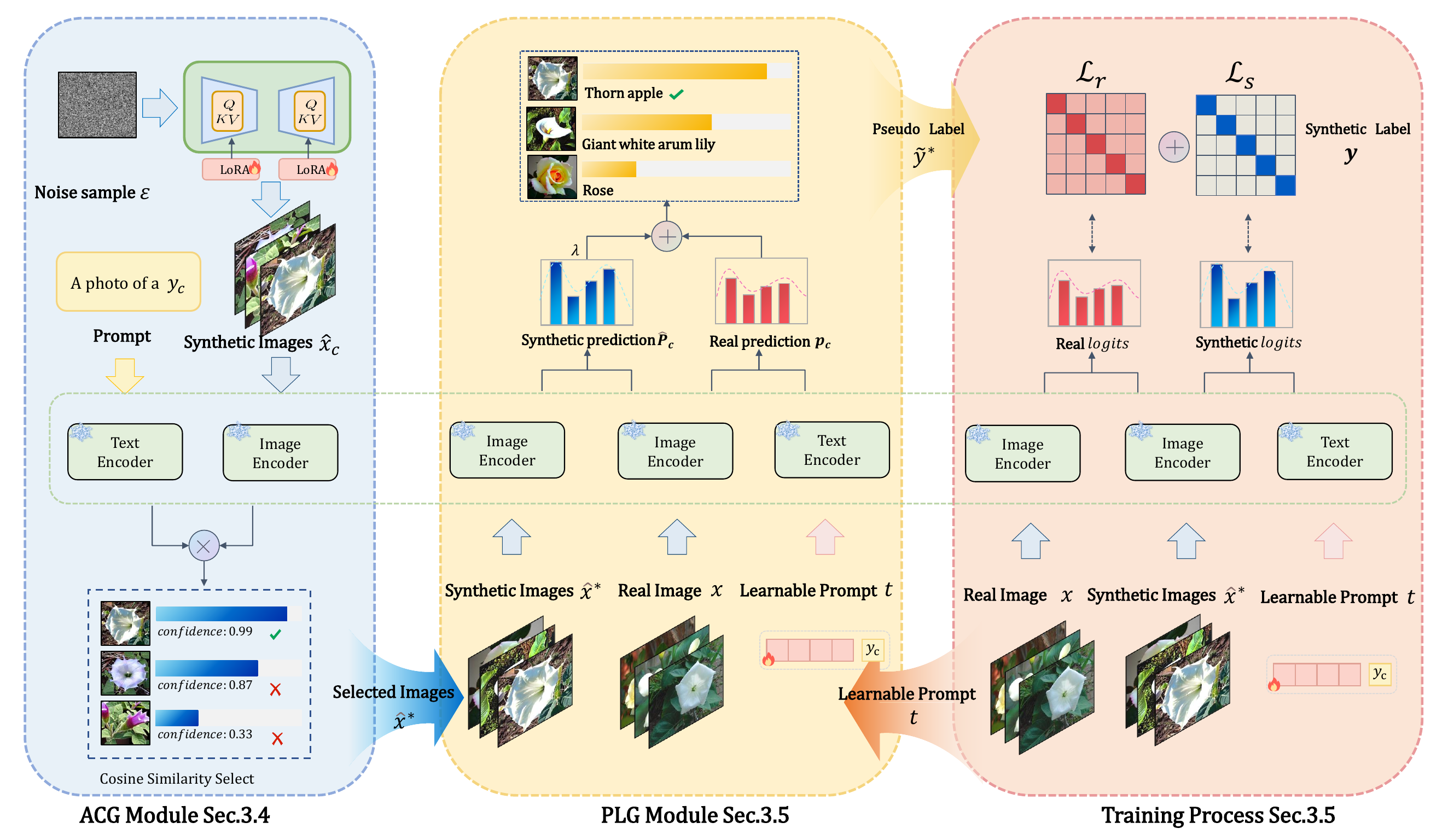}
\caption{{
Overview of our proposed AiR, which consists of ACG (Auxiliary Classifier Generation) Module and PLG (Pseudo Label Generation) Module.
ACG generates synthetic samples using the LoRA fine-tuned SD model and selects the representative samples by cosine similarity to build auxiliary classifier. PLG generates  pseudo-labels by fusing the predictions of the auxiliary classifier built on synthetic images and the text classifier.
The overall loss for training prompt consists of the classification loss of real images with pseudo-labels $\mathcal{L}_{r}$ and classification loss of synthetic images with corresponding categories $\mathcal{L}_{s}$, where ours can learn visual and textual prompt.}
}
\vspace{-10pt}
\label{fig:method}
\end{figure*}

%% file: Sections/2_related.tex
\section{Related Work}
\label{sec: rw}

\subsection{Prompt Learning}
\label{ssec:pl}
Prompt learning is a technique that enhances the practical application of large pre-trained models like CLIP \cite{radford2021learning} and GPT \cite{brown2020language, radford2019language} by enabling improved performance across various downstream tasks through tailored text or visual prompts. 
For instance, CoOp~\cite{zhou2022learning} optimizes continuous prompt vectors instead of relying on discrete ones, thereby avoiding the suboptimal performance caused by hand-designed prompts. 
CoCoOp~\cite{zhou2022conditional} addresses CoOp's generalization challenges by incorporating Meta-Net, which balances performance between base and novel classes.
When only unlabeled data is available, some work is dedicated to unsupervised learning to accomplish prompt learning.
UPL~\cite{huang2022unsupervised} using CLIP’s~\cite{radford2021learning} zero-shot capabilities, followed by confidence refinement to generate pseudo-labels with higher confidence for unsupervised learning.
IFPL and GRIP~\cite{menghini2023enhancing}, proposed `Iterative Refinement of Few-pseudo labels' and `Grow and Refine Iteratively Pseudolabels' respectively, in terms of how to utilize pseudo-labelled samples.
To alleviate the issue of hard pseudo labels, CPL \cite{zhang2024candidate}, proposes a
Candidate Pseudolabel Learning method, which progressively generates refined candidate pseudo labels by both intra- and inter-instance label selection. \gdd{Different from these pseudo-labeling \cite{menghini2023enhancing, zhang2024candidate, huang2022unsupervised} approaches, which typically rely on text classifiers to generate pseudo labels. 
We propose the AiR framework, which uses synthetic images as a bridge to expand text-image-pair classification with image-image-pair classification, further improving the model's discriminative performance from a richer visual perspective.}
\subsection{Training with Synthetic Data}
The diffusion model (DM) has shown significant effectiveness in various synthetic tasks, including image \cite{ho2020denoising, song2020denoising}, video \cite{bar2022text2live}, audio \cite{kong2020diffwave}, and text \cite{li2022diffusion}.
Concurrently, DM has made notable advancements in image manipulation and restoration tasks, such as image editing \cite{avrahami2022blended}, inpainting \cite{lugmayr2022repaint}, and deblurring \cite{whang2022deblurring}.
In the field of prompt learning, exploration has also been undertaken. 
For example, DiffTPT \cite{feng2023diverse} generates diverse augmented images with diffusion models while faithfully preserving the key semantics.
DataDream \cite{kim2024datadream}, a novel few-shot method that adapts Stable Diffusion to generate better in-distribution images for downstream training and is also proven effective in the field of prompt learning.
However, most of the aforementioned methods primarily focus on leveraging diffusion models (DM) for data augmentation or generating more realistic images, without employing DM as a bridge to connect the unsupervised text modality with the image modality. 
Furthermore, the application of DM within unsupervised prompt learning (UPL) remains largely unexplored.

%% file: Sections/3_method.tex
\section{Method}
\label{sec:method}

\subsection{Preliminaries}
\label{ssec:pre}
\noindent\textbf{Revisiting Unsupervised Prompt Tuning.}
We provide a brief introduction to the prevailing unsupervised prompt learning (UPL) method.
In the UPL setting, we have access only to unlabeled data $X_U = \{x_i\}_{i=1}^N$ composed of $N$ instances with target classes name $\{y_1, ..., y_C \}$.
\gdd{Suppose a VLM with learnable prompt parameters $t$. 
 Besides, in the CLIP zero-shot setting, ``a photo of a" is a usually used prompt, denoted as $t_{zs}$.} Each class name $\{y_c\}_{c=1}^C$ is concatenated with a text prompt to formulate a text input ``a photo of a [CLS]", which is represented as $w_{c}=\{\rr{t_{zs}}, y_c\}$ and can be fed into the text encoder $G$ in VLM to yield a text embedding $g_c = G(w_{c})$. Concurrently, given an instance $x$ in $X_U$, we can embed it with image encoder $F$ of CLIP to produce a visual embedding $f = F(x)$.
The prediction probabilities over classes are given as follows:
\begin{equation}
p_c = \frac{\text{exp}(\text{sim}({f}, {g}_c) / \tau)}{\sum_{c^{\prime}=1}^C \text{exp}(\text{sim}({f}, {g}_{c^{\prime}}) / \tau)},
\label{eq:zc}
\end{equation}
where $\tau$ represents the temperature coefficient.
Besides, $\text{sim}(\cdot)$ refers to a metric function for measuring the similarity between the input image and the text, which is realized with cosine similarity.
Mainstream UPL methods such as FPL \cite{huang2022unsupervised}, GRIP \cite{menghini2023enhancing} employ CLIP's zero-shot capability to assign a pseudo-label $\tilde{y}$ to each sample $x$ as follows:
\begin{equation}
\tilde{y} = \underset{c}{\text{argmax}}p_c,
\label{eq:pl}
\end{equation}
typically, one can use a top-$K$ pseudo labeling approach, where the top-$K$ most confident examples per class are used as pseudo-labeled data.
Then, the prompts $t$ can be optimized via back-propagating the following loss:
\begin{equation}
\underset{t}{\text{min}}\mathcal{L}_{ce} = \underset{t}{\text{min}}\mathbb{E}_{({x},\tilde{y})} \left[-\text{log}p(\tilde{y}|{x})\right].
\label{eq:ce}
\end{equation}
However, due to the limited expressive power of purely semantic information, the accuracy of pseudo-labels generated by text-only classifiers remains sub-optimal.

\subsection{Overview}
This work aims to improve the pseudo-labeling accuracy in UPL from a novel view. Specifically, we extend text-image-pair classification to image-image-pair classification by generating synthetic samples for each category, enhancing the classifier’s visual perception with the Diffusion Model (DM), which serves as a bridge. The overall workflow of our method can be divided into three steps: 
\textcircled{1} 
Finetune DM with LoRA.
Considering the need to ensure the diversity of the synthetic samples while making the synthetic samples more consistent with the domain distribution of the current downstream dataset, we use the LoRA strategy to finetune the DM(see Sec. \ref{ssec:fd}).
\textcircled{2} 
\textbf{ACG} (Auxiliary Classifier Generation) Module: After finetuning with LoRA, we generate synthetic images through the DM with descriptions of the dataset and class name.
Then, we obtain the auxiliary classifier with cosine similarity and select a representative sample for each class.
(see Sec. \ref{ssec:sd}).
\textcircled{3}
\textbf{PLG} (Pseudo Label Generation) Module: Based on the synthetic image, we expand text-image-pair prediction with image-image-pair prediction, further improving the model's discriminative performance from a richer visual perspective and obtaining a more comprehensive pseudo label (see Sec. \ref{ssec:lw}). 

\subsection{Finetune DM with LoRA}
\label{ssec:fd}
While the DM can generate a diverse range of images through text-to-image generation, these images do not always align with the dataset of the downstream task. 
This may introduce spurious classes and style mismatches, leading to domain inconsistency and degraded prompt tuning performance.
Thus, it is necessary to balance the augmented images' data diversity and domain consistency.
The most straightforward approach would be to fine-tune the DM using labeled datasets from downstream tasks; however, this method is computationally expensive, and labeled data is unavailable. \gdd{To this end, we adopt CLIP's zero-shot generated pseudo-labeled data $\{x, \tilde{y}\}$ in Sec.~\ref{ssec:pre} to fine-tune the DM with LoRA \cite{hu2022lora}, where we employ Stable Diffusion (SD) and the details about the LoRA is deferred to Appendix.}
The original SD model parameters $\theta$ are kept frozen and only the LoRA weights $\delta$ are trained with the objective function:
\begin{equation}
\underset{\delta}{\text{min}}\mathcal{L}_{lora} \!=\! \underset{\delta}{\text{min}}\mathbb{E}_{\rr{({x},\tilde{y})}, \epsilon \sim \mathcal{N}(0, 1)}([||\epsilon \!-\!\epsilon_{\theta, \delta}(x,t_{data},\tilde{y})||_2^2],
\label{eq:lora}
\end{equation}
where $\epsilon \sim \mathcal{N}(0, 1)$ denotes the sampled noise, $\epsilon_{\theta, \delta}(\cdot)$ denotes the noise generated by SD model. \gdd{In addition to the training image $x$ and its pseudo-label $\tilde{y}$ generated by zero-shot CLIP, the description of the dataset $t_{data}$ is also the input of the SD model, to ensure that the generation of samples is more in line with the current dataset. Taking the Flowers102 dataset as the example, $t_{data}$ is ``Flowers:'' and $\tilde{y}$ may be ``Thorn apple'' or any class else.} In this way, we ensure both the diversity of the synthesized samples and domain consistency to the current dataset domain. For subsequent synthetic images, we used the LoRA-updated SD model for generation.

\subsection{ACG Module}
\label{ssec:sd}
After fine-tuning with LoRA, SD is presented to generate diverse and domain-consistency data.
For each class in the UPL task, we can generate synthetic data with the SD model $S(\cdot)$ as:
\begin{equation}
\rr{\hat{X}_{c}^M = S(\{t_{data}, y_c\}, \epsilon)},
\label{eq:ds}
\end{equation}
where $\hat{X}_{c}^M = \{\hat{x}_{c}^i\}_{i=1}^M$ represents the generated synthetic images of class $y_c$ and $M$ denotes the number of images generated for each class. 
Considering the synthetic image needs to be used as the auxiliary classifier, it is essential to select the representative sample from $\hat{X}_c^M$ that best aligns with the semantic information of the current category. Following \cite{feng2023diverse}, we compute the cosine similarity between the zero-shot CLIP text embedding $g_c$ and each synthetic image in $\hat{X}_c^M$. Using Eq. \ref{eq:zc}, we then select the synthetic sample with the highest confidence $p_c$ for each class $c$ and achieve the following synthetic dataset:
\begin{equation}
\hat{X}^* = \{\hat{x}_1^*, \hat{x}_2^*, ..., \hat{x}_C^* \}.
\label{eq:xf}
\end{equation}
\gdd{We now build the bridge between the text class name and the representative synthetic dataset. Here, $\hat{x}_c^*$ is the most representative synthetic image for the $c$-th class and it contains the ample visual perceptual discrimination information, exactly what the textual class name used in zero-shot CLIP lacks. Therefore, it is natural to extract the visual embedding of $\hat{X}^*$ with the vision encoder $F$ and build the auxiliary classifier based on the visual prototypes.}



\subsection{PLG Module \& Learning with Synthetic Image}
\label{ssec:lw}
\gdd{Through the above auxiliary classifier, we can correct the original prediction probabilities in Eq. \ref{eq:zc} by the zero-shot CLIP as:}
\begin{equation}
\begin{aligned}
{p}_{c}^* &= p_c + \lambda * \hat{p}_c,
\\ \hat{p}_c &= \frac{\text{exp}(\text{sim}(f, F(\hat{x}_c^*)) / \tau)}{\sum_{c^{\prime}=1}^C \text{exp}(\text{sim}(f,F(\hat{x}^*_{c^{\prime}})) / \tau)},
\label{eq:pi}
\end{aligned}
\end{equation}
where $\lambda \geq 0$ serves as a hyper-parameter of these two type predictions. By combining the two predictions, the model's discrimination of the current image contains both semantic and visual perceptual information.
Depending on the discrimination result in Eq. \ref{eq:pi}, we are able to obtain a more comprehensive pseudo label $\tilde{y}^*$ for each sample $x$ as $\tilde{y}^* = \underset{c}{\text{argmax}}{p}_{c}^*$, 
then, we can use these pseudo-labeled data to optimize the prompt $t$, and the training loss can be expressed as:
\begin{equation}
\underset{t}{\text{min}}\mathcal{L}_{r} = \underset{t}{\text{min}}\mathcal{L}_{ce}(x, \tilde{y}^*).
\label{eq:lr}
\end{equation}

Considering that the synthetic samples generated by stable diffusion in Eq.~\ref{eq:ds} also extend the diversity information of the current category while being labeled counterparts, we participate in the synthetic samples in the training process as well as:
\begin{equation}
\underset{t}{\text{min}}\mathcal{L}_{s} = \underset{t}{\text{min}}\mathcal{L}_{ce}(\hat{x}, y).
\label{eq:ls}
\end{equation}
In conclusion, the training objective of our method can be formally expressed as:
\begin{equation}
\underset{t}{\text{min}}\mathcal{L} = \underset{t}{\text{min}}(\mathcal{L}_{r} + \beta * \mathcal{L}_{s}),
\label{eq:l}
\end{equation}
where $\beta \geq 0$ represents the coefficient of equilibrium between the two loss functions.
After each iteration, the optimized prompt $t$ is sent back to the PLG module and processed through Eq.~\ref{eq:pi} to generate a more accurate pseudo label.
This iterative process facilitates the continuous refinement and optimization of pseudolabels. 
Thus, we not only obtain high-quality pseudo-labeled samples that are classified from semantic and visual perspectives with richer visual perception information but also a diversity of high-quality synthetic samples to aid in training, thus improving the unsupervised prompt learning process.

\subsection{The Overall Training Process}
\label{ssec:to}

In this subsection, we outline the overall training process of our method, AiR. 
To balance data diversity and domain consistency in the augmented images, we begin by fine-tuning SD with LoRA. 
Next, synthetic images $\hat{X}$ are generated through SD, and we select the sample best aligned with the semantic information based on cosine similarity. 
The chosen synthetic images $\hat{X}^*$ then act as auxiliary classifiers, in combination with text classifiers, to assign pseudo-labels $\tilde{y}^*$ to the unlabeled samples. 
Finally, the pseudo-labeled data $(x, \tilde{y}^*)$ and synthetic data $(\hat{x}, y)$ serve as training samples to calculate the loss and iteratively optimize the prompt. 
The algorithm of AiR is summarized in the Appendix.


%% file: Sections/4_experiments.tex
\section{Experiments}
\label{sec:exp}

\input{Tables/text_visual_prompt}

\subsection{Benchmark Setting}
\noindent\textbf{Datasets.}
We conduct a comprehensive evaluation of our method on five diverse classification datasets: DTD \cite{cimpoi2014describing}, EuroSAT \cite{helber2019eurosat}, FGVC \cite{maji2013fine}, Flowers102 \cite{nilsback2008automated}, and RESISC45 \cite{cheng2017remote}. 
These datasets span a wide array of topics, including general object, scene, and fine-grained classification, along with specialized tasks such as texture recognition and satellite imagery analysis.

\noindent\textbf{Learning Paradigms.}
To comprehensively evaluate the effectiveness of our method, we investigate three common scenarios involving unlabeled data: Unsupervised Learning (UL), Semi-Supervised Learning (SSL), and Transductive Zero-Shot Learning (TRZSL). 
Detailed explanations of each paradigm and our approach to labeled data are provided in the Appendix.

\noindent\textbf{Baselines.} 
To evaluate the effectiveness of the training strategies outlined in Sec. \ref{sec:method}, we compare ours against CLIP's zero-shot performance using default prompts such as `a photo of a [CLASS]'.
We also benchmark against standard supervised prompt-tuning methods that use only labeled data: CoOp \cite{zhou2022learning} for textual prompts and VPT \cite{jia2022visual} for visual prompts.
Additionally, we compare our approach, AiR, with three existing unsupervised prompt-tuning methods: Few Pseudolabels (FPL) \cite{menghini2023enhancing}, Grow and Refine Iteratively Pseudolabels (GRIP) \cite{menghini2023enhancing}, and CPL \cite{zhang2024candidate}.

\noindent\textbf{Evaluation metrics.}
We evaluate each method by calculating test set accuracy and averaging results over three runs for consistency. 
In the case of Transductive Zero-Shot Learning (TRZSL), we report the harmonic mean of accuracies for seen and unseen classes to address potential performance imbalances between them.

\noindent\textbf{Implementation details.}
In all experiments, spanning various datasets and learning strategies, we use ViT-B/32 as the vision backbone. 
We use Stable Diffusion v1-4 as our synthetic data generation model.
For both visual and textual prompt tuning, the prefix size is set to $16$. 
We employ SGD as the optimizer and train for 150 epochs, with an initial learning rate of 0.0001 during the first 5 warmup epochs, after which it is adjusted to $l$ and decayed according to a cosine annealing schedule. 
For both textual and visual prompt tuning, $l$ is set to 0.1.
The number of iterations is set to 10. 
For FPL, the number of pseudo-labels per class is fixed at 16, following \cite{menghini2023enhancing}. 
In GRIP and CPL, the number of pseudo-labels is treated as a tunable hyperparameter following \cite{zhang2024candidate}.

\subsection{Comparison with SOTA Methods}
\label{sec:cw}

To validate the effectiveness of our proposed AiR, we employ CPL~\cite{zhang2024candidate} as a baseline and conduct comparative experiments using three learning paradigms: SSL, UL, and TRZSL on five distinct datasets, benchmarking against contemporary SOTA methods. 
To demonstrate the generalizability of our approach, we evaluate AiR in both text prompt learning and visual prompt learning frameworks.
As shown in Table~\ref{tab:tvp} our approach consistently achieves SOTA results across various datasets and learning paradigms. 
In UL, AiR enhances performance by an average of approximately 2.6\% in text prompt learning and 2.7\% in visual prompt learning. 
In both SSL and TRZSL, AiR achieves performance improvements ranging from 1\% to 3\%. These results indicate that AiR consistently boosts the classification performance of prompt learning models across different learning paradigms. 
This underscores the robustness and generalizability of our approach, demonstrating that augmenting discriminative richness effectively enhances unsupervised prompt learning performance in diverse scenarios.
We also conducted experiments on the large-scale dataset and cross-dataset settings, please refer to the Appendix.


\subsection{Ablation Studies}
\noindent\textbf{Effect of Synthetic Prediction and Synthetic Loss.}
We conduct ablation studies on three datasets using the UL and text prompt paradigm to demonstrate the individual contributions of each core component in our AiR method. 
In Table~\ref{tab:abl}, $\hat{p}_{c}$ denotes the inclusion of discriminative results from synthetic samples in Eq. \ref{eq:pi}, $\mathcal{L}_{s}$ indicates whether these synthetic samples contribute to the loss calculation in Eq. \ref{eq:ls}, and $\mathcal{L}_{r}$ indicates whether real samples contribute to the loss calculation in Eq. \ref{eq:lr}. 
To isolate our approach’s impact, we use GRIP \cite{menghini2023enhancing} as the baseline and report model accuracy under each condition. 
When real samples loss function $\mathcal{L}_{r}$ is employed,  using only $\hat{p}_{c}$ yields a 1.7\%-5.4\% improvement, indicating that enriching visual perception discriminative information has a substantial effect on model accuracy. 
Similarly, training with synthetic samples from the SD model together with real samples enhances performance by 1.5\%-4.8\%, showing that sample diversity and fidelity provide valuable information to the unsupervised model, helping it adapt prompts better suited to downstream tasks. 
However, given that our AiR model leverages synthetic samples generated by the SD model for the unsupervised prompt learning task, an important question arises: if only synthetic samples are used for training, is it feasible to fully address the unsupervised learning task? 
As the result shows, the model's performance declines significantly when no real samples are included in the training, for instance, a decrease of $-10.1\%$ in Flowers102 and $-12.3\%$ in RESISC45. 
These results indicate that relying solely on synthetic samples is insufficient for unsupervised learning. Although synthetic samples exhibit sufficient diversity and real-world fidelity, they still cannot fully replace the information provided by real samples.
When $\hat{p}_{c}$, $\mathcal{L}_{s}$, and $\mathcal{L}_{r}$ are combined, the model achieves optimal performance, highlighting that the enhanced visual perception and sample diversity effects are complementary rather than redundant, yielding a compounded improvement. 
We also conducted experiments on the effect of different generation models, please refer to the Appendix for more details.


\noindent\textbf{Effect of LoRA.}
To assess the necessity of LoRA for training SD models, we evaluate the accuracy of SD models with and without LoRA on three datasets (RESISC45, Flowers102, EuroSAT, and DTD) in the unsupervised paradigm, with CPL as the baseline. As illustrated in Table \ref{tab:lnl}, synthetic samples generated with LoRA fine-tuning provide our AiR model with a 3\%-8\% accuracy gain compared to the case without LoRA in the same training environment. 
This result highlights that while SD can produce diverse and high-quality synthetic samples, alignment with the dataset’s domain is crucial, making LoRA an indispensable component of our AiR model.
We also conduct ablation experiments on cosine similarity for the sample selected, please refer to the Appendix for more details.

\input{Tables/ablation}
\subsection{Exploratory Analysis}


\noindent\textbf{Sensitivity to the Number of Synthetic Samples.}
To investigate the impact of varying numbers of synthetic samples on model performance, we conducted experiments assessing changes in model accuracy with $0$–$150$ synthetic samples per class on the RESISC45 and EuroSAT datasets, respectively.
As shown in Fig.~\ref{fig:nu}, the trend line illustrates how model accuracy shifts with the increasing number of synthetic samples. 
We observe that model performance gradually improves as the number of synthetic samples rises, especially when increasing from $60$ to $90$ samples: EuroSAT accuracy improves by $3\%$, and RESISC45 accuracy by $1\%$. 
However, this improvement is not indefinite; beyond $90$ synthetic images, performance gains slow, and after surpassing $120$ samples, the trend stabilizes. 
This suggests that synthetic samples indeed enrich discriminative information for unsupervised prompt learning and in turn enable noticeable improvement of the model's performance, though there is a limit to this benefit. 
Based on these observations, we set the number of synthetic samples at $120$ across all learning scenarios in our AiR framework.
More experiments about the hyper-parameter $\lambda$ in Eq.~\ref{eq:pi} and $\beta$ in Eq.~\ref{eq:l} are provided in the Appendix.

\input{Figures/num}


\input{Tables/lora_vs_no_lora}

\noindent\textbf{Pseudo Label Accuracy.}
To demonstrate the extent to which the auxiliary classifier, constructed with synthetic samples, improves the pseudo-labeling quality of the model's original text classifier.
We use the original prompt ``a photo of a" and compare the accuracy of pseudo-labeling for the top 50 maximum confidence samples. 
We employ synthetic samples generated by our AiR model as a reference across three different datasets.
As shown in Fig. \ref{fig:pla}, leveraging synthetic sample-assisted classification within our AiR model results in at least a 3\% improvement in the original text classification accuracy, with a 6\% gain on RESISC45. 
This indicates that augmenting the discriminative richness through enhanced visual perceptual information significantly improves pseudo-labeling accuracy, which in turn boosts the performance of unsupervised prompt learning models.
More experiments about pseudo-label accuracy are provided in the Appendix.

\input{Figures/pla}

\input{Figures/cam}

\subsection{Qualitative Analysis}
\noindent\textbf{CAM of Different Discrimination Outcomes.}
To show the effectiveness of synthetic image embeddings intuitively, we use CLIP \cite{radford2021learning} as a baseline and perform CAM (Class Activation Mapping) \cite{zhou2016learning} visualization on two datasets. 
As shown in Fig.~\ref{fig:cam}, we observe that text embeddings sometimes focus on class-irrelevant regions, such as airports in the \emph{Airplane} class, seawater in \emph{Bridge}, and grass in \emph{Blackberry Lily}. 
Additionally, text embeddings may fail to capture all relevant areas, such as part of a \emph{Beach} or a flower petal in \emph{Mexican Aster}. 
When using synthetic samples for discrimination, similar issues arise, but the regions of attention differ. 
For instance, while text embeddings overly focus on airports in \emph{Airplane}, synthetic samples pay less attention to class-relevant regions of the airplane. 
Similarly, text embeddings focus on grass in \emph{Blackberry Lily}, while synthetic samples shift focus to the location of stamens.
This suggests a fundamental difference in the type of information emphasized by different discriminative modalities for the same sample. 
Moreover, when combining the discriminative information from both text and synthetic images, we find that the fused results provide a more complete focus on class-relevant regions while minimizing the influence of class-irrelevant areas. 
For example, the fused attention highlights larger regions of the beach in \emph{Beach} and more complete flowers in \emph{Fire Lily} and \emph{Mexican Aster}.
This result confirms why our AiR method works because when the visual perceptual information embedded in the synthetic samples is used to enhance the richness of the discrimination, it helps the model to focus more on class-relevant regions and suppress activation in class-irrelevant areas, thus resulting in an improvement in model performance.

\input{Figures/tsne}

\noindent\textbf{t-SNE of Different Discrimination Outcomes.}
In the previous section, we demonstrated the differences in image activation regions across various discriminative methods using CAM. 
In this section, we examine the spatial distribution of synthetic image embeddings, text embeddings, and sample embeddings through t-SNE visualization.
As shown in Fig.~\ref{fig:tsne}, circular dots in different colors represent distinct image classes in Flowers102, triangles indicate text embeddings for each class, squares represent synthetic image embeddings, and pentagrams denote the embeddings after fusing text and synthetic image embeddings as outlined in Sec.~\ref{ssec:lw}.
From the figure, it is apparent that due to modality differences, text embeddings (triangles) collapse into a compact region that is only loosely aligned with the corresponding class's test samples, rather than overlapping with them. 
Conversely, the synthetic sample embeddings (squares) align directly with data clusters for each class, appearing more suitable as classifiers compared to the text embeddings.
However, the observed tight clustering of text embeddings in two-dimensional t-SNE space is primarily due to modality discrepancies, which may not reflect their true spatial distribution in higher-dimensional space. 
Thus, we cannot conclude outright that synthetic image embeddings are categorically more suitable for classification.
Nevertheless, by fusing the embeddings (pentagrams), we see a clear tendency for text embeddings to move toward their corresponding test sample classes. 
This shift suggests that augmenting discriminative information effectively calibrates the embeddings, aligning them more closely with the correct test samples and thus improving the model’s classification accuracy across different classes.
More experiments on t-SNE are provided in the appendix.

\noindent\textbf{Fidelity of generated samples $w$ and $w/o$ LoRA.}
We provide examples of images generated by the SD model with and without LoRA to visually demonstrate the importance of LoRA in our AiR model.
As shown in Fig.~\ref{fig:syn}, images generated without LoRA bear a general resemblance to the real images, yet differ significantly upon closer inspection. 
For instance, the petal shapes in \emph{Canterbury Bells}, the viewpoint in \emph{Airplane}, and the texture details in \emph{Blotchy} reveal substantial mismatches; without LoRA, the SD model can only approximate the overall class, lacking finer, more precise details.
In contrast, with LoRA applied, the generated images exhibit more accuracy in form and detail: the petal shapes in \emph{Canterbury Bells} closely match the real image, the airplane's viewpoint in \emph{Airplane} shifts to a top view, and the texture in \emph{Blotchy} appears smoother and more aligned.
These improvements demonstrate that without LoRA, synthetic samples lack sufficient detailed expression to real samples and cannot effectively serve as classification embeddings to augment discriminative information. 
This underscores the necessity of LoRA in our AiR model.

\input{Figures/syt}

%% file: Tables/text_visual_prompt.tex
\begin{table*}[!t]
\centering

\resizebox{\textwidth}{!}{

\begin{tabular}{ l | c  | c c c c c c c c c  c c c c c c c c c }
	\toprule
        \textcolor{blue}{TP} &   & & RESISC45 & & & FGVCAircraft & &  & Flowers102  & & & EuroSAT & & & DTD &  & & Average  &\\ 
      \hline
       Method & Year  & SSL & UL & TRZSL & SSL & UL & TRZSL & SSL & UL & TRZSL & SSL & UL & TRZSL & SSL & UL & TRZSL & SSL & UL & TRZSL \\
      \hline
       CLIP\cite{radford2021learning} & 2021  & 54.5 & 54.5 & 54.5 & 17.6 & 17.6  & 17.9 & 63.7 & 63.7 & 63.4 & 32.8 & 32.8 & 30.5 & 43.2 & 43.2 & 43.4 & 42.3 & 42.3 & 41.9 \\
       CoOp\cite{zhou2022learning}  & 2022  & 58.5 & - & 63.4 & 14.9  & - & 21.7 & 76.8 & - & 63.2 &59.5 & - & 49.6 & 37.1 & - & 46.3 & 49.3 & - & 48.8\\
      FPL\cite{menghini2023enhancing}  & 2023  & 68.1 & 63.0 & 72.1  & 20.0 & 16.6  & 17.5 & 75.9 & 65.6 & 80.9 &62.0 & 48.9 & 53.7 & 37.1 & 44.9 & 46.3 & 52.6  &  47.8 & 54.0\\
       GRIP\cite{menghini2023enhancing}  & 2023  & 74.1 & 70.6 & 81.1 & 17.0 & 15.2 & 26.1 & 83.6 & 69.8 & 86.8 & 58.6 & 57.2 & 92.3 & 56.0 & 46.0 & 65.3 & 57.8 & 51.7 & 70.3\\
       CPL\cite{zhang2024candidate} & 2024  & 80.9 & 77.3 & 85.8 & 22.4  & 18.3  & 30.8  & 89.6 & 72.9 & 87.3 &75.4 & 67.2 & 93.7 & 61.2 & 51.9 & 68.0 & 65.9 & 57.5 & 73.1 \\
      \rowcolor{gray!15}
      Ours & -  & \textbf{82.5} & \textbf{79.9} & \textbf{87.5} & \textbf{24.2} &\textbf{19.6} & \textbf{33.2} & \textbf{91.2} & \textbf{74.3} & \textbf{92.3} &\textbf{76.4} & \textbf{71.4} & \textbf{93.9} & \textbf{62.2} & \textbf{55.7} & \textbf{69.9} & \textbf{67.3} & \textbf{60.1} & \textbf{75.4} \\
      \hline
      \textcolor{red}{VP} &   & & RESISC45 & & & FGVCAircraft & &  & Flowers102  & & & EuroSAT & & & DTD &  & & Average  &\\ 
      \hline
       CLIP\cite{radford2021learning} & 2021 & 54.5 & 54.5 & 54.5 & 17.6 & 17.6  & 17.9  & 63.7 & 63.7 & 63.4 & 32.8 & 32.8 & 30.5 & 43.2 & 43.2 & 43.4 &42.3  & 42.3 & 41.9 \\
       VPT\cite{jia2022visual} & 2022 & 60.8 & - & 67.1 & 17.7  & - & 26.6 & 63.7 & - & 64.7 & 47.1 & - & 62.2 & 36.4 & - & 44.1 & 45.1 & - & 52.9 \\
       FPL\cite{menghini2023enhancing} & 2023  & 65.1 & 62.2 & 67.8  & 20.1 & 18.2 & 16.2 & 67.0 & 65.5 & 71.9 & 52.4 & 48.7 & 68.6 & 47.6 & 47.6  & 52.4  & 50.4 & 48.4 & 55.3 \\
       GRIP\cite{menghini2023enhancing} & 2023  & 71.2 & 68.4 & 82.1  & 19.4 & 17.5  & 26.4 & 67.9 & 63.0 & 77.1 & 63.4 & 63.6 & 96.9 & 54.5 & 50.5 & 62.7 & 55.2 & 52.6 & 68.4 \\
      CPL\cite{zhang2024candidate} & 2024  & 78.4 & 72.9 & 86.6 & 20.5  & 18.2  & 30.2   & 73.5 & 67.2 & 80.1 &72.0 & 68.9 & 98.3 & 58.7 & 53.4 & 65.3 & 60.6 & 56.1 & 72.1 \\
      \rowcolor{gray!15}
        Ours & -  & \textbf{81.1} & \textbf{79.3} & \textbf{87.8} & \textbf{23.7} &\textbf{19.3} & \textbf{31.2} & \textbf{78.5} & \textbf{69.2} & \textbf{81.9} &\textbf{73.5} & \textbf{71.6} & \textbf{98.5} & \textbf{59.6} & \textbf{54.6} & \textbf{66.4} & \textbf{63.2} & \textbf{58.8} & \textbf{73.2} \\

	\bottomrule
\end{tabular}}
\caption{Comparison results of top-1 test accuracy($\%$) on five benchmarks.
\textcolor{blue}{TP} and \textcolor{red}{VP} represent applying text prompts and visual prompts as tuning strategies, respectively.
The average accuracy on all datasets is shown on the right.
The best results are in \textbf{bold}.
}
\label{tab:tvp}
\vspace{-10pt}
\end{table*}

%% file: Tables/ablation.tex
\begin{table}[!t]
\centering

\resizebox{.32\textwidth}{!}{
\begin{tabular}{c c c c c c}
\toprule
 $\mathcal{L}_{r}$  & $\hat{p}_{c}$ & $\mathcal{L}_{s}$  & RESICS45   & Flowers102 & DTD \\ 
\hline
$\checkmark$ & &  & 70.6 & 69.8 & 46.0  \\
 &  & $\checkmark$ & 58.3 & 59.7 & 41.4 \\
$\checkmark$   &$\checkmark$  &   & 72.3 & 71.6  & 51.4 \\ 
$\checkmark$ & & $\checkmark$  & 72.9 & 71.3 & 50.8  \\ 
\rowcolor{gray!15}
$\checkmark$ & $\checkmark$  & $\checkmark$ & \textbf{73.6} & \textbf{71.9} & \textbf{52.3} \\ 
\bottomrule
\end{tabular}
}

\caption{Accuracy scores ($\%$) on RESISC45, Flowers102, and DTD are reported to verify the effects of different strategies.
$\hat{p}_{c}$ represents the auxiliary classifier from synthetic samples, $\mathcal{L}_s$ and $\mathcal{L}_r$ indicate the training loss about the synthetic samples and real samples, respectively.
}

\label{tab:abl}
\vspace{-20pt}
\end{table}

%% file: Figures/num.tex
\begin{figure}
\centering

\includegraphics[width=.9\linewidth]{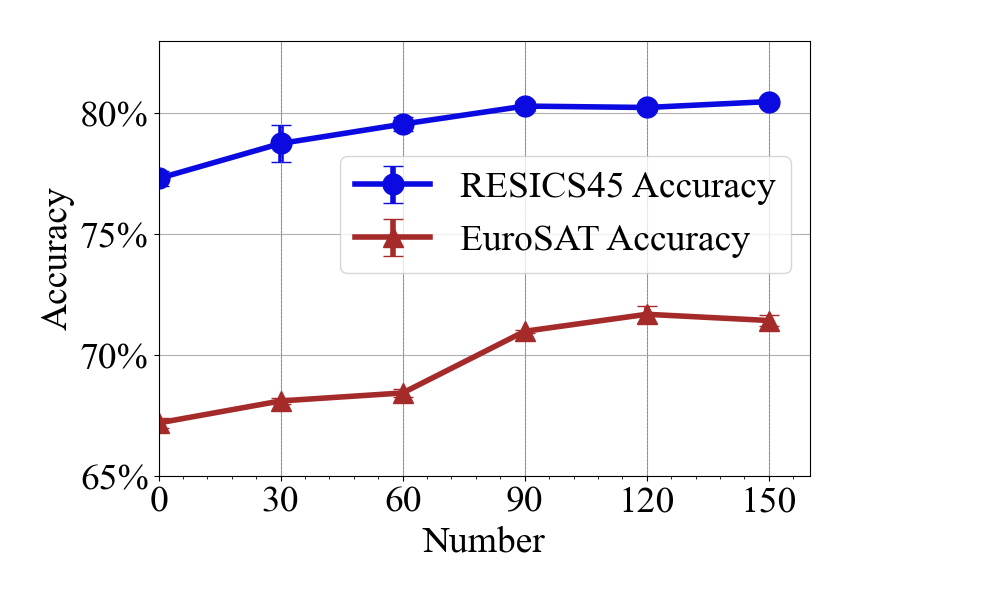}
\caption{
Comparison of top-1 test accuracy ($\%$) across varying numbers of synthetic samples in unsupervised learning on the RESISC45 and EuroSAT datasets.
}
\vspace{-10pt}
\label{fig:nu}
\end{figure}

%% file: Tables/lora_vs_no_lora.tex
\begin{table}[!t]
\centering

\resizebox{.39\textwidth}{!}{
\begin{tabular}{l l l l l}
	\toprule
      LoRA & RESISC45 & Flowers102 &EuroSAT &  DTD  \\ 
      \hline
      $w/o$  & 76.5 & 66.3 & 66.2   & 51.6 \\
      \rowcolor{gray!15}
      $w$  & \inc{\textbf{79.9}}{3.4}  & \inc{\textbf{74.3}}{8.0} & \inc{\textbf{71.4}}{5.2} & \inc{\textbf{55.7}}{4.1} \\
      \bottomrule
\end{tabular}}
\caption{Comparison results of top-1 test accuracy ($\%$) on unsupervised learning with or without LoRA fine-tuning.
Note that `w/o' denotes the synthetic sample generated without LoRA.
}
\label{tab:lnl}
\vspace{-15pt}
\end{table}

%% file: Figures/pla.tex
\begin{figure}
\centering

\includegraphics[width=.8\linewidth]{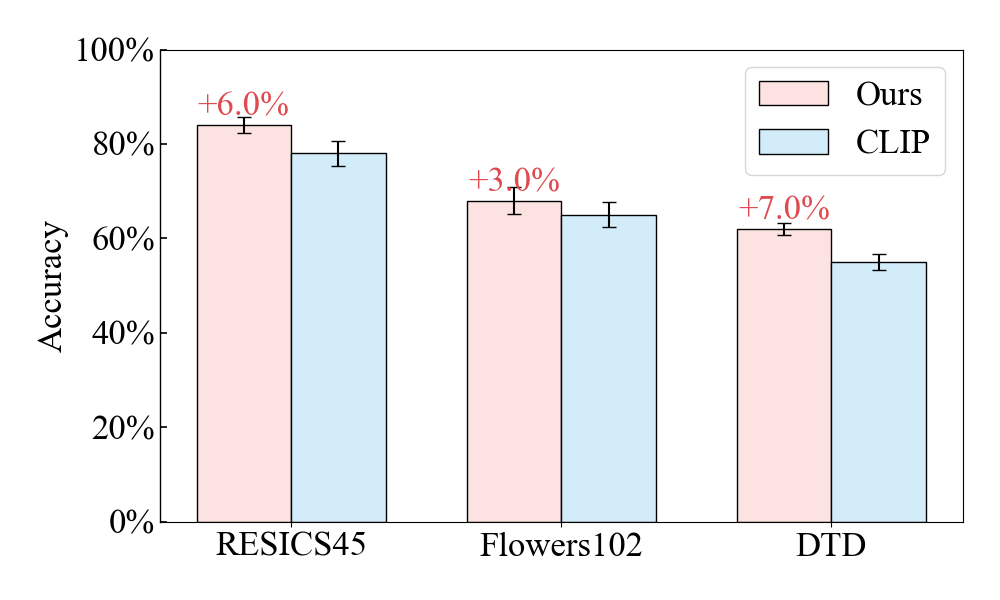}
\caption{
Comparison of top-1 test accuracy ($\%$) of pseudo labels in unsupervised learning. 
`Ours' denotes the use of our approach AiR, while `CLIP' indicates the use of CLIP's text encoder.
}
\vspace{-15pt}
\label{fig:pla}
\end{figure}

%% file: Figures/cam.tex
\begin{figure}
\centering

\includegraphics[width=.9\linewidth]{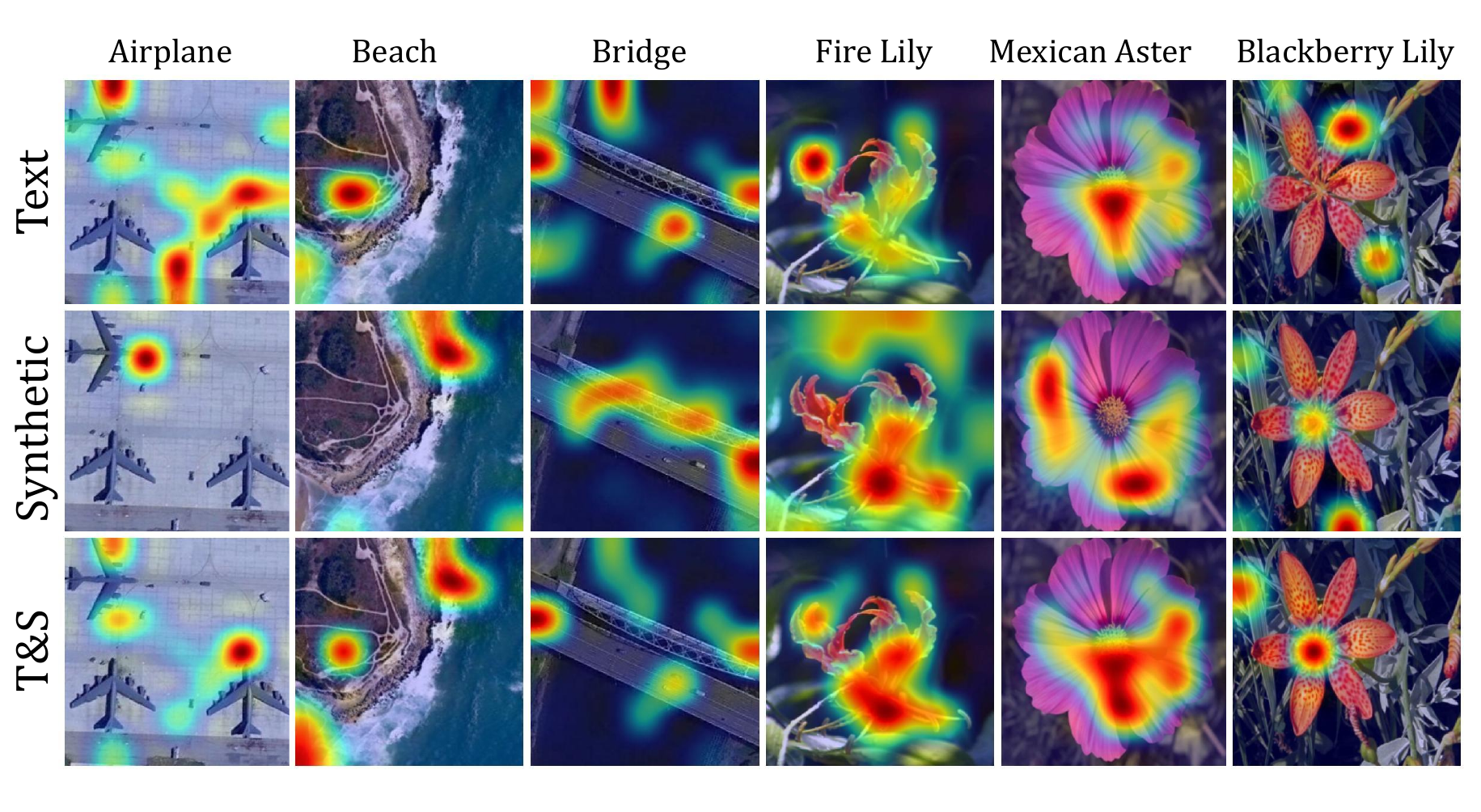}
\caption{
Visualization of the activation regions for the text classifier and synthetic-sample-based auxiliary classifier using CAM. 
`T $\&$ S' represents the activation after merging the outputs of both classifiers.
}
\vspace{-15pt}
\label{fig:cam}
\end{figure}

%% file: Figures/tsne.tex
\begin{figure}
\centering

\includegraphics[width=.9\linewidth]{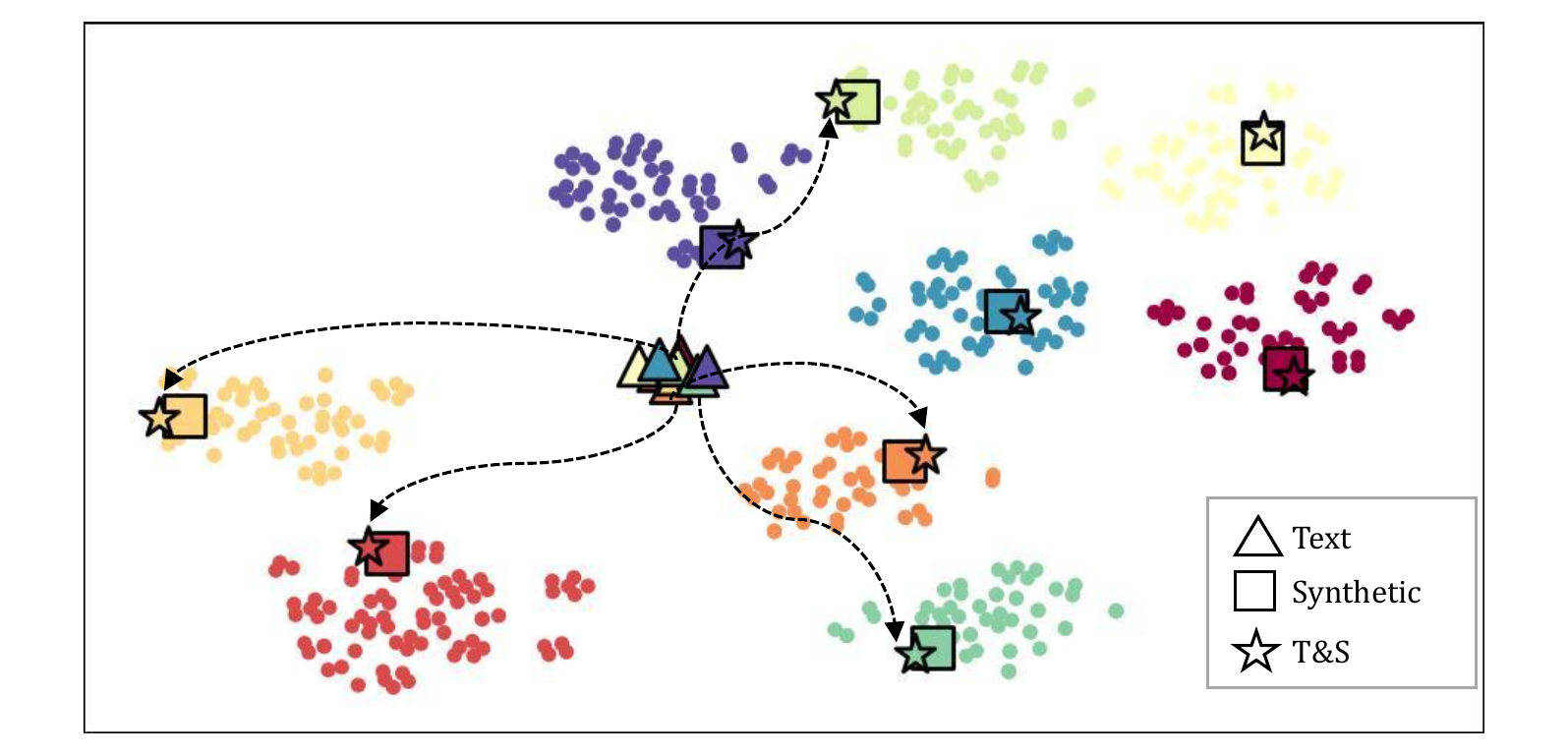}
\caption{
Visualization of the spatial distribution of synthetic image, text, and sample embeddings
with t-SNE, respectively.
Circular dots in different colors represent distinct image classes in Flowers102, triangles indicate text embeddings for each class, squares represent synthetic image embeddings, and pentagrams denote the embeddings after fusing text and synthetic image embeddings.
}
\vspace{-20pt}
\label{fig:tsne}
\end{figure}

%% file: Figures/syt.tex
\begin{figure}
\centering

\includegraphics[width=.9\linewidth]{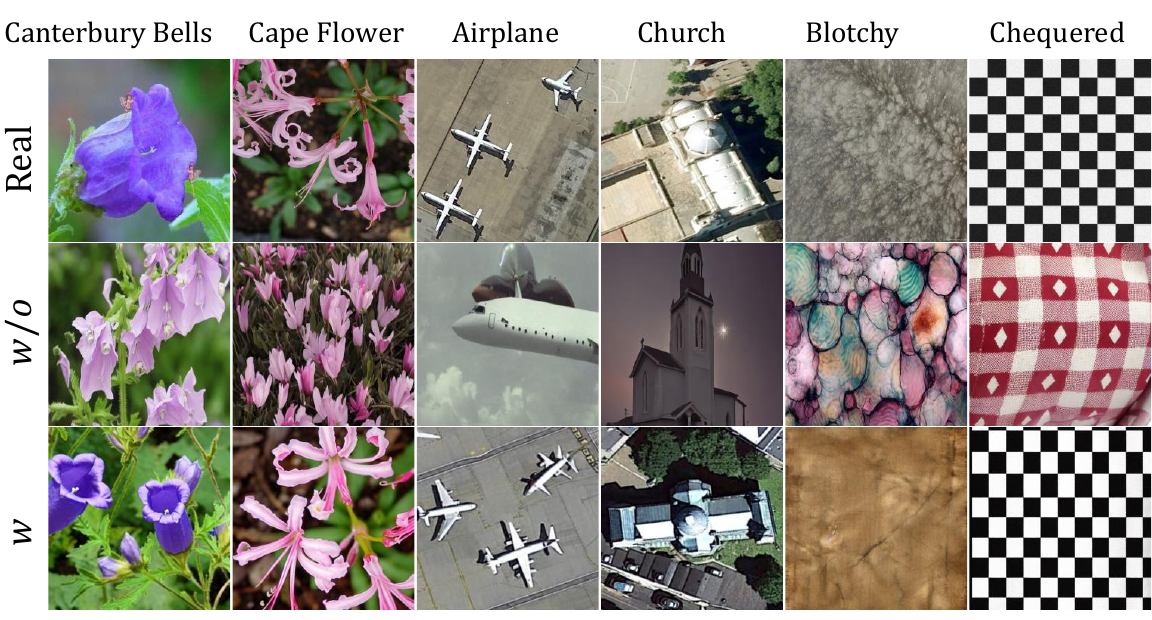}
\caption{
Visualization of real images and synthetic images generated by the SD model, with and without fine-tuning via LoRA.
}
\vspace{-15pt}
\label{fig:syn}
\end{figure}

%% file: Sections/5_conclusion.tex
\section{Conclusion}
In this paper, we introduce AiR, a novel auxiliary classification approach designed to leverage synthetic samples in building an auxiliary classifier. 
In this way, we can capture richer visual variations while preserving core semantics, effectively bridging text-image-pair classification to image-image-pair classification. 
Extensive experiments across five benchmark datasets and three learning paradigms validate the effectiveness of our approach. 
We hope this work inspires new approaches to reducing dependence on labeled data when adapting pre-trained vision-language models like CLIP to novel tasks.

%% file: Sections/7_Ack.tex
\paragraph{Acknowledgements.}
 This work of Dandan Guo was supported by the National Natural Science Foundation of China (NSFC) under Grant 62306125.
 This work of Yi Chang was supported in part by the National Key R\&D Program of China under Grant 2023YFF0905400, in part by the National Natural Science Foundation of China under Grant U2341229. 
 This work of Fan Tang was supported in part by the Beijing Science and Technology Plan Project under no. Z231100005923033, and in part by Beijing Natural Science Foundation under no. L221013.